# Evolutionary Reinforcement Learning via Cooperative Coevolutionary Negatively Correlated Search

Hu Zhang; Peng Yang; Yanglong Yu; Mingjia Li; Ke Tang
Southern University of Science and Technology

## Abstract

Evolutionary algorithms (EAs) have been successfully applied to optimize the policies for Reinforcement Learning (RL) tasks due to their exploration ability. The recently proposed Negatively Correlated Search (NCS) provides a distinct parallel exploration search behavior and is expected to facilitate RL more effectively. Considering that the commonly adopted neural policies usually involves millions of parameters to be optimized, the direct application of NCS to RL may face a great challenge of the large-scale search space. To address this issue, this paper presents an NCS-friendly Cooperative Coevolution (CC) framework to scale-up NCS while largely preserving its parallel exploration search behavior. The issue of traditional CC that can deteriorate NCS is also discussed. Empirical studies on 10 popular Atari games show that the proposed method can significantly outperform three state-of-the-art deep RL methods with 50% less computational time by effectively exploring a 1.7 million-dimensional search space.

## Section I Introduction

Reinforcement Learning (RL) is an attracting machine learning problem that actively learns the decision-making policy through environmental interactions with delayed and noisy rewards, without consuming labeled training data like traditional supervised learning [1]. In RL, an agent sequentially observes the environment state and takes an action to the environment suggested by the policy, a reward is then returned (from the environment) to the agent reflecting how it interacts with the environment. By iteratively conducting the above procedure, RL is targeted to learn the policy that can maximize the long-term reward [2].

For this purpose, the policy is expected to accurately react to the observed environment states. The arising Deep Reinforcement Learning (DRL) methods provide the possibility by employing deep neural networks as the policy [3]. DRL builds policy directly on the observed raw data, offering seamless interfaces to the environment, and facilitates the state-action decision-making with powerful non-linear inference ability for many successful real-world applications [4]-[6]. On the other hand, due to the non-uniform distribution of rewards, it is important to explore the action space to enlarge the coverage of the policy on the state space [7]. This requires to exploratively optimize the parameters of the policy, e.g., the connection weights of deep networks. Traditional gradient-based methods [8][9] are not always effective due to the lacks of exploration ability.

Recently, Evolutionary Algorithms (EAs) have been successfully applied to DRL by searching for the optimal policy in a population-based manner [10]-[12]. The population-based nature of EAs not only provides the urgent exploration ability to RL, but also offers other bonuses such as parallel acceleration [13], noisy-resistance [14], and compatibility of training non-differentiable policies

(e.g., trees [15]). Due to those advantages, the resultant Evolutionary Reinforcement Learning (ERL) has drawn increasing research popularity from both academia and industries.

Actually, the balance between exploration and exploitation has been extensively studied in the EA community for decades [16]. Basically, it is widely acknowledged that the diversity of the population is an important indicator to guide the exploration. Rich volume of works have been proposed to effectively measure the diversity of the current population and adjust it accordingly to generate the next population. Until recently, [17] discusses that the diversity of the current population may not be closely related to the generation of the diversified next population, and proposes to directly measure the diversity of the next population. Based on this idea, the presented Negatively Correlated Search (NCS) is able to capture the on-going interactions between successive iterations and introduce manageable diversity to the generation of the next population, successfully forming a fully parallel exploration search behavior where each individual in the population is managed to visit a different region of the search space with high enough objective quality. Due to its effectiveness, NCS has been successfully applied to several types of real-world applications [18]-[21]. This paper aims to apply NCS into ERL to offer a parallel exploration ability for policy training.

Note that training a deep neural network needs to optimize commonly millions or more connection weights. Such huge numbers of decision variables usually impose great challenges to EAs (e.g., NCS) [13]. Generally, to solve such large-scale problems, the Cooperative Coevolution (CC) framework is often beneficial for scaling up EAs by dividing the decision variables into multiple exclusive groups and addressing the smaller sub-problems (i.e., groups of decision variables) with EAs respectively [22]. One major difficulty of applying CC lies in the evaluation of partial solutions in each sub-problem with respect to the original objective function. Existing CC methods mostly complement all partial solutions in a sub-problem with the same complementary vector [22], and then evaluate the complemented solutions accordingly. This strategy is mostly effective for traditional EAs whose population tend to converge and the individuals in the population are somewhat similar at the later stage of the search. However, due to the parallel exploration behaviors, the population of NCS tend to diverse. In this case, sharing the same complementary vector among all partial solutions may lead to significantly incorrect evaluations of some partial solutions, and thus fail the parallel exploration. To address this issue, a specified CC framework is devised to scale up NCS without compromising its exploration ability, resulting in the proposed Cooperative Coevolutionary Negatively Correlated Search (CCNCS).

CCNCS is applied to train a neural policy with 1.7 million connection weights on 10 popular Atari games to verify its effectiveness on RL problems. Empirical results show that CCNCS can outperform the state-of-the-art DRL methods (including both gradient-based and EA-based policy training methods) by scoring significantly higher scores within half less computational time. Furthermore, it has been shown that CCNCS can obtain scores in some games with very sparse reward distributions, while the state-of-the-art methods cannot break the tie. This should also be attributed to the well-preserved strong exploration ability while scaling up NCS with CC. The reminder of this paper is organized as follows. Section II briefly introduces the preliminaries of ERL, NCS, and CC. Section III details the proposed CCNCS method. Section IV presents the empirical studies of CCNCS on Atari games. The conclusions are drawn in Section V.

## Section II Preliminaries

This section briefly provides three aspects of preliminaries for the following sections.

**Section II.A Evolutionary Reinforcement Learning**

ERL specifies the class of RL approaches that using EAs to directly optimize the parameters of the policy [10], e.g., deep neural networks in this paper. The general flowchart of ERL can be seen in Fig.1. Initially, each individual of an EA is represented as a vector of all the connection weights of the policy. The training phase is divided into multiple epochs. At each epoch, the agent starts from the beginning of a game and takes actions suggested by a policy model (i.e., an individual of EA) with respect to the environment states. After a game (i.e., an epoch) has been finished, the reward of the policy will be returned back to the agent as well as EA. Then EA takes the rewards of all candidate policies in the current population to generate the new population of policies for the next epoch. The above procedure repeats until the training budget runs out, and the best policy model at the final epoch will be output.

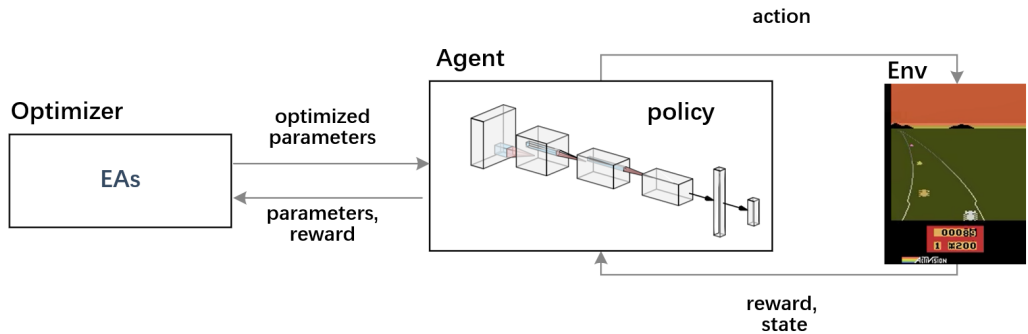

Fig. 1 The flowchart of Evolutionary Reinforcement Learning

ERL enjoys multiple merits offered by EAs. First, the population-based search mechanism of EAs can be exploited to improve both the exploration and parallel acceleration of ERL [11]. Second, the search direction of EAs can be guided by the relative order of the individuals in terms of their objective qualities, which makes ERL less sensitive to the evaluation noise [14]. Lastly, EAs can be used to optimize non-differentiable models (e.g., trees [15]), thus ERL is not restricted to train neural policy.

**Section II.B Negatively Correlated Search**

NCS is a recently proposed EA that features in its parallel exploration behavior [17]. Technically, NCS regards the population of $\lambda$ individuals as $\lambda$ exclusive search processes. Each search process is evolved with a traditional EA to maximize the objective quality of each individual. Meanwhile, the diversity among different search processes are explicitly measured and maximized to improve the exploration ability. Different from most existing EAs that measure the diversity as the differences among the individuals of the current population, NCS measures the diversity as the differences among individuals of the next population. By maximizing the diversity of the unsampled next population, it is straightforward to diversify the search processes in a controllable manner. As a result, each search process can visit a different region of the search space while preserving high objective quality, finally forming a parallel exploration search behavior (see Fig.2 for illustration).

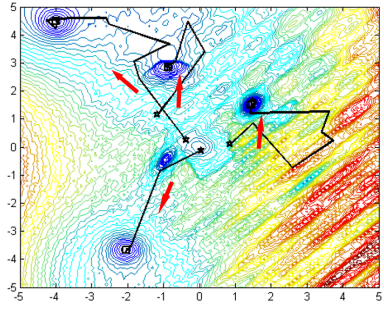

Fig.2 The parallel exploration search pattern of NCS is illustrated on a 2-D multi-modal function, i.e., F19 in CEC'2005 continuous optimization functions benchmark [28], where 4 search processes (i.e., 4 individuals) explore different regions of the search space and successfully locate 4 optima.

Take the instantiated NCS-C [17] as an example. At each iteration, each $i$-th search process preserves a parent individual $\mathbf{x}_i$, and generates an offspring individual $\mathbf{x}'_i$ from the Gaussian distribution $\mathcal{N}(\mathbf{x}_i, \mathbf{\Sigma}_i)$. The Gaussian distribution basically describes how the $i$-th search process will generate the next individual. Thus, if the Gaussian distributions of different search processes can be explicitly forced to be different, they are highly likely generate diversified individuals. To realize this idea, three steps are conducted in NCS-C. First, the difference among distributions is measured by the Bhattacharyya distance, which considers the differences of both mean vectors and covariance matrices and is able to quantify the "overlap" of pairwise Gaussian distributions. Second, the diversity of the next population is modeled as the differences of all search processes and is maximized parallel with respect to each search process. Specifically, for each search process, either the parent distribution $\mathcal{N}(\mathbf{x}_i, \mathbf{\Sigma}_i)$ or the offspring distribution $\mathcal{N}(\mathbf{x}'_i, \mathbf{\Sigma}'_i)$ is selected for the next iteration to maximize the whole diversity. Third, while selecting either the parent or offspring, their objective qualities are also considered. Thus, a balance between the exploration (by maximizing the diversity) and the exploitation (by maximizing the objective quality) can be readily achieved by a trade-off parameter, which in turn is managed to control the distributions of search processes to be negatively correlated for parallel exploration. The pseudo code of NCS-C is given in Algorithm 1 for illustration. For more details of NCS, please refer to [17].

| **Algorithm 1** NCS-C |
| --- |
| 1. **Initialize** $\lambda$ solutions $\mathbf{x}_i$ randomly, $\mathbf{x}_i, i = 1, ..., \lambda$. |
| 2. **Evaluate** the $\lambda$ solutions with respect to the objective function $f$. |
| 3. **Repeat** until stop criteria are met: |
| 4.     **For** $i$=1 to $\lambda$ |
| 5.         **Generate** an offspring solution $\mathbf{x}'_i$ from the distribution $p_i \sim \mathcal{N}(\mathbf{x}_i, \mathbf{\Sigma}_i)$. |
| 6.         **Calculate** $f(\mathbf{x}'_i)$, $d(p_i)$ and $d(p'_i)$.     // $d$ denotes the diversity |
| 7.         **If** $f(\mathbf{x}'_i) + \varphi \cdot d(p'_i) > f(\mathbf{x}_i) + \varphi \cdot d(p_i)$     // $\varphi$ denotes the trade-off parameter |
| 8.            **Update** $(\mathbf{x}_i, \mathbf{\Sigma}_i)$ with $(\mathbf{x}'_i, \mathbf{\Sigma}'_i)$. |
| 9.         **Update** $\mathbf{\Sigma}_i$ according to certain EA rules. |
| 10. **Output** the best solution ever found with respect to $f$. |

**Section II.C Cooperative Coevolution**

EAs basically work by randomized sampling in the search space, which means their performance usually drop heavily when the search space enlarges significantly. To solve such large-scale problems with EAs, a commonly adopted method is CC [22]. CC follows the classic divide-and-conquer methodology that first divides the decision variables into multiple small-scale groups, and optimizes each group separately with an EA. Ideally, each EA only needs to tackle a small-scale sub-problem while the original large-scale problem can still be effectively solved [22].

The major difficulty of applying the divide-and-conquer idea to EAs lies in how to evaluate the partial solutions in each sub-problem. To solve this problem, CC first complements the low-dimensional partial solutions to the dimensionality of the original objective function, and then evaluates them accordingly. However, different complementary vectors to a same partial solution can result in different objective qualities. Thus, rich amount of research efforts has been devoted to studying the complementing methods so as to improve the accuracy of the evaluation of partial solutions [23]. There are also a plenty of approaches that have been proposed to the grouping of decision variables, aiming to make the sub-problems independent from each other so that even random complementary vectors can lead to accurate evaluations [24]. Also note that other methods that do not need to complement partial solutions also exist [13].

Despite of all the technical details, most CC methods basically complement all the partial solutions in a sub-problem with the same complementary vector [22]. This is reasonable in many cases since keeping the complementary vector the same for all partial solutions can largely reduce the noise of calculating their relative order that guides the search direction of many EAs. However, it will be shown in the next section that such strategy does not work well for NCS due to its parallel exploration search behavior. For illustration, the framework of classic CC is briefly given as follows.

| **Algorithm 2** Classic Cooperative Coevolution |
|---|
| 1. **Initialize** $\lambda$ solutions $\mathbf{x}_i$ randomly, $\mathbf{x}_i, i = 1, \dots, \lambda$. |
| 2. **Repeat** until stop criteria are met: |
| 3.     **Divide** the $D$-dimensional problem into $M$ sub-problems. |
| 4.     **For** $j$=1 to $M$ |
| 5.         **Evolve** all the partial solutions $x_{i,j}$ for one iteration with an EA, $i = 1, \dots, \lambda$. |
| 6.         **For** $i = 1$ to $\lambda$ |
| 7.             **Complement** each partial solution $x_{i,j}$ with the same vector $v_j$ as $[x_{i,j}, v_j]$. |
| 8.             **Evaluate** each complemented solution $f([x_{i,j}, v_j])$ as the qualities of $x_{i,j}$. |
| 9. **Output** the best complemented solution ever found with respect to $f$. |

# Section III Cooperative Coevolutionary Negatively Correlated Search

This section first discusses that the traditional CC framework is not suitable for NCS, and an NCS-friendly CC framework is proposed accordingly. The resultant CCNCS by incorporating NCS into the variant CC framework is also described in detail.

**Section III.A The NCS-Friendly CC Framework**

As mentioned above, traditional CC complements all the partial solutions of a sub-problem with the same complementary vector and then evaluates the complemented solutions with respect to the objective function. This is useful for most EAs as their partial solutions in each sub-problem often tend to be similar at the convergence stage of the search, where the population highly likely locates at the same sub-space of a basin of attraction (i.e., they will converge to the same optimum). In this case, the differences of the evaluation qualities among all partial solutions can be made marginal by complementing them with the same complementary vector. By "marginal", we mean this optimum will be found by this population very likely, no matter by which specific individual.

Distinctly, NCS explicitly asks each search process to be negatively correlated, where the partial solutions in each sub-problem tend to be very different from each other at the later stage of the search and the population may locate multiple different basins of attraction. In this case, complementing all the partial solutions with the same complementary vector is problematic. For example, Fig.3 illustrates a 2-D search space with two optima located at the left-upper corner and the right-lower corner. Suppose NCS has obtained two partial solutions in the first sub-problem, denoted as $x_{1,1}$ and $x_{2,1}$, respectively. If they are complemented with $x_{1,2}$, it is clear that $f(x_{1,1}, x_{1,2})$ is better than $f(x_{2,1}, x_{1,2})$ and the left-upper optimum will be found. If they are complemented with $x_{2,2}$, the right-lower optimum will be found. In either case, the traditional CC will deteriorate the parallel exploration ability of NCS as either of the two optima will be missed out. As the iterations go on, the population of NCS will finally get converged to the partial solution who best fits the complementary vector, and thus fails to form the featured parallel exploration. One can also complement the partial solutions with multiple complementary vectors (e.g., both $x_{1,2}$ and $x_{2,2}$ in Fig.3) to fit each promising partial solution with redundancy. However, it is computationally expensive as multiple times of objective function evaluations will be consumed.

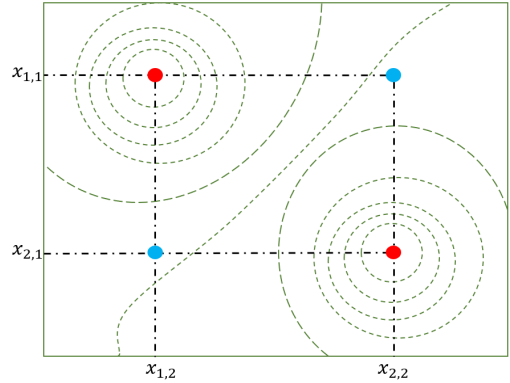

Fig.3 Traditional CC that complements two partial solutions of a sub-problem, i.e., $x_{1,1}$ and $x_{2,1}$, with the same vector, i.e., either $x_{1,2}$ or $x_{2,2}$, is not suitable for NCS alike search methods as either optimum (marked as red solid circles) will be missed.

In this paper, we propose to complement each partial solution of NCS with a different complementary vector to maximally preserve the parallel exploration search behavior. The only difference between the NCS-friendly CC and traditional CC lies in the step 7 of Algorithm 2. Specifically, in each $j$-th sub-problem, each $i$-th partial solution $x_{i,j}$ will be complemented with a distinct complementary vector $v_{i,j}$ in the new CC framework, rather than the same complementary vector $v_j$ as for traditional CC. This helps to avoid the situation in Fig.3 where one complementary vector cannot fit diversified promising partial solutions. For the context of the $i$-th complementary vector $v_{i,j}$, we simply set it as $v_{i,j} = \mathbf{x}_j/\mathbf{x}_{i,j}$. That is, each $i$-th complementary vector in the $j$-th sub-problem is calculated as the quotient set of $\mathbf{x}_j$ with respect to $x_{i,j}$, and is composed of the current $i$-th partial solution in other sub-problems. As the complementary vector is evolved together with the partial solution, i.e., $[x_{i,j}, v_{i,j}] = \mathbf{x}_j$, this complementing strategy should also be able to well characterize the quality of each $x_{i,j}$. This strategy actually exploits the parallel exploration feature itself that different partial solutions will locate the sub-space of different basins of attraction. Also note this strategy does not cost extra computational objective function evaluations as each partial solution is still evaluated once.

**Section III.B The Details of CCNCS**

In this section, NCS-C is incorporated into the variant CC framework to obtain the proposed CCNCS and the detailed pseudo code is illustrated in Algorithm 3.

| **Algorithm 3** CCNCS |
|---|
| 1. **Initialize** $\lambda$ individuals $\mathbf{x}_i$ randomly, $i = 1, \ldots, \lambda$. |
| 2. **Repeat** until stop criteria are met: |
| 3.     **Divide** the $D$-dimensional problem into $M$ sub-problems. |
| 4.     **For** $j$=1 to $M$ |
| 5.       **For** $i$=1 to $\lambda$ |
| 6.         **Generate** an offspring solution $x'_{i,j}$ from the distribution $p_{i,j} \sim \mathcal{N}(x_{i,j}, \Sigma_{i,j})$. |
| 7.         **Calculate** the diversity of both parent and offspring as $d(p_{i,j})$ and $d(p'_{i,j})$. |
| 8.         **Complement** $x_{i,j}$ and $x'_{i,j}$ as $[x_{i,j}, v_{i,j}]$ and $[x'_{i,j}, v_{i,j}]$, respectively. |
| 9.         **Evaluate** $f([x_{i,j}, v_{i,j}])$ and $f([x'_{i,j}, v_{i,j}])$ as the qualities of $x_{i,j}$ and $x'_{i,j}$. |
| 10.         **If** $f([x'_{i,j}, v_{i,j}]) + \varphi \cdot d(p'_{i,j}) > f([x_{i,j}, v_{i,j}]) + \varphi \cdot d(p_{i,j})$ |
| 11.           **Update** $(x_{i,j}, \Sigma_{i,j})$ with $(x'_{i,j}, \Sigma'_{i,j})$. |
| 12.         **Update** $\Sigma_{i,j}$ according to NCS rules. |
| 13. **Output** the best complemented solution ever found with respect to $f$. |

CCNCS first randomly initialize $\lambda$ individuals (i.e., search processes) in the original $D$-dimensional search space. At each iteration of the main loop, the large-scale $D$ decision variables are first decomposed in to $M$ groups. In the literature, there are lots of well-established decomposition methods [25]. For simplicity, the random grouping with varied values of $M$ is employed [26]. To be brief, when executing the step 3, the value of $M$ is first randomly selected from a fixed pool with candidate values, and the $D$ decision variables are randomly divided into $M$ groups. Then NCS is run for optimizing each sub-problem (steps 5-12). As NCS conducts $\lambda$ search processes in parallel, the detailed search operators can be executed with respect to each partial solution. Specifically, for the $i$-th partial solution in the $j$-th sub-problem $x_{i,j}$, its offspring $x'_{i,j}$ is first generated from the distribution $p_{i,j} \sim \mathcal{N}(x_{i,j}, \Sigma_{i,j})$. Then the Bhattacharyya distance between the parent distribution $p_{i,j} \sim \mathcal{N}(x_{i,j}, \Sigma_{i,j})$ to the distributions of other search processes is calculated and denoted as $d(p_{i,j})$. Similarly, the Bhattacharyya distance of the offspring distribution $p'_{i,j} \sim \mathcal{N}(x'_{i,j}, \Sigma'_{i,j})$ to the distributions of other search processes is calculated and denoted as $d(p'_{i,j})$. To measure the qualities of $x_{i,j}$ and $x'_{i,j}$, they are first respectively complemented as $[x_{i,j}, v_{i,j}]$ and $[x'_{i,j}, v_{i,j}]$, and evaluated with respect to the objective function $f$. In fact, $f([x_{i,j}, v_{i,j}])$ can be obtained from the previous iteration, thus it will not cost function evaluations except for the first iteration. Then by considering both the objective qualities and the diversity, i.e., $f([x'_{i,j}, v_{i,j}]) + \varphi \cdot d(p'_{i,j})$ and $f([x_{i,j}, v_{i,j}]) + \varphi \cdot d(p_{i,j})$, either the parent partial solution or its offspring is selected to the next iteration. Accordingly, the distribution of either the parent or the offspring should also be passed to the next iteration as it describes the search behavior of the search process. After that, the distribution should be adjusted based on the feedback of the search process. Basically, the distribution mainly concerns the mean vector and covariance matrix. Between them, the mean vector is actually identical to the partial solution and should not be adjusted. The covariance matrix can be adjusted via various rules (e.g., the 1/5 successful rule adopted in [17]). After the main loop terminates, the best complemented solution ever found will be output as the final solution of CCNCS.

## Section IV Empirical Studies

This section verifies the effectiveness of CCNCS on the popular RL problems, i.e., playing Atari Games. First, the Atari Games and the CCNCS-based ERL method are described. After that, the experimental protocol including the experimental settings and the compared state-of-the-art algorithms are briefly introduced. Finally, empirical results are analyzed to discuss the effectiveness of CCNCS.

**Section IV.A CCNES for Playing Atari Games**

Atari 2600 is a set of video games that have been popular for benchmarking RL approaches [10]. Atari games successfully cover different types of difficult tasks, such as obstacle avoidance (e.g., Freeway and Enduro), shooting (Beamrider, and SpaceInvaders), maze (e.g., Alien and Venture), balls (e.g., Pong), Mario alike (e.g., Montezuma's Revenge), Sports (e.g Bowling and DoubleDunk), and other types. Those types of games provide significantly different environmental settings and interaction ways for the agent, and thus are able to test the RL approaches with different tasks in terms of maximizing the long-term reward. Another feature of Atari games is that the reward is delayed for actions as it is only returned after a game is finished. Furthermore, the environment of Atari games can involve considerable randomness that the rewards may be noisy. For the underlying empirical studies, the mentioned-above 10 games are selected to assess the performance of CCNCS.

When playing Atari games, CCNCS fully follows the ERL flowchart depicted in Fig.1. Specifically, each individual is represented as a vector of all connection weights of the neural policy model (i.e., one individual of CCNCS represents one candidate policy), and is evolved based on Algorithm 3. Each epoch of training is equivalent to each iteration of CCNCS, and one game playing with a candidate policy is actually one objective function evaluation of an individual of CCNCS. In this paper, the policy suggested in [8] is adopted, which is composed of three convolution layers and two full connection layers. The architecture of the network can be seen in Table I, which involves nearly 1.7 million connection weights to be optimized. With the convolution layers, the policy is able to directly process the raw pixel data from the videos and no more efforts are needed for preprocessing the environmental states.

**Table I  The Network Architecture of the Agent**

|  | Input | Output | Kernel Size | Stride | #filters | activation |
|---|---|---|---|---|---|---|
| **Conv1** | 4x84x84 | 32x20x20 | 8x8 | 4 | 32 | ReLU |
| **Conv2** | 32x20x20 | 64x9x9 | 4x4 | 2 | 64 | ReLU |
| **Conv3** | 64x9x9 | 64x7x7 | 3x3 | 1 | 64 | ReLU |
| **Fc1** | 64x7x7 | 512 | - | - | - | ReLU |
| **Fc2** | 512 | #Actions | - | - | - | - |

**Section IV.B Experimental Protocol**

Three state-of-the-art DRL approaches are employed as the compared algorithms, i.e., A3C [8], PPO [9], and CES [10]. Among them, A3C and PPO are popular gradient-based methods that train the network with the traditional back-propagation. CES is a recently proposed ERL that utilizes the canonical Evolution Strategies to directly optimize the policy network. By comparing with those three methods, it suffices to assess how CCNCS performs against the state-of-the-arts. Besides, in order to show how CC facilitates NCS, we directly apply NCS to Atari games as the baseline.

All the above four algorithms are targeted to train the same policy network with CCNCS. Each algorithm terminates the training phase in a game when the total budget runs out, and the best policy network at the final epoch will be returned for testing. The quality of the final policy is measured with the testing score, i.e., averaged score of 200 repeated games playing without the frame limitations. Given that the environment is randomly initialized, each run will be re-run for three times, i.e., each algorithm will obtain three testing scores on each game. The total time budget is set as the total game frames that each training phase is allowed to consume. For three ERL methods (i.e., CES, NCS, and CCNCS), the total game frames are set to 100 million. For A3C and PPO, as it works quite differently with back-propagation, it is unfair to set the same total game frames with the ERL methods. On this basis, we counted the game frames consumed by the well-established CES and A3C on the same hardware configurations with the same given computational run time in the same games. It has been found that the ratio of the consumed game frames between A3C and CES is about 2.5. As a result, the total game frames are set to 40 million for A3C and PPO for fairness. To discretize the games for agent's actions execution and states acquiring, the skipping frame is set to 4. In this case, in each training phase, the agent is allowed to take 25 million actions for derivate-free methods and 10 million actions for gradient-based methods.

For A3C, PPO, and CES, the hyper-parameters suggested in their original papers are directly employed for the comparisons. For CCNCS, the population size $\lambda$ is set to 6. The covariance of the Gaussian distribution is initialized as 0.2 for each dimensionality. The number of sub-problems $M$ is randomly picked from the set of (2,3,4) at each iteration. The other hyper-parameters of CCNCS follow the original settings of NCS in [17]. To keep it simple, NCS shares the same hyper-parameters with CCNCS, expect that it does not need to divide $M$ sub-problems.

**Section IV.C Results and Analysis**

Three repeated testing scores of each algorithm on 8 games are shown in Table II, where the best averaged score of the three testing scores in each game is marked in bold. It can be seen that CCNCS generally performs the best among all compared algorithms and NCS also obtains competitive results in most cases. Specifically, for Alien, Beamrider, Freeway, Montezuma's Revenge, Pong, SpaceInvader, Venture, Bowling and DoubleDunk, CCNCS can significantly outperform the 3 state-of-the-art methods within the same time budget. For Enduro, CCNCS can still perform better than A3C and CES while performs less satisfactorily than PPO. To further express the effectiveness of CCNCS, we run it with 50 million game frames which is 50% less than the time budget of the other algorithms. The results are listed in the rightmost column of Table II. It is clearly seen that the advantages still exist that CCNCS can gain as much as twice scores than the state-of-the-art methods in most of the tested games within half less time.

Among all tested games, Montezuma's Revenge is almost the hardest one as traditional well-established methods (including the three compared ones) can hardly gain any score, unless human experience is incorporated to educate the policy [28]. This is mainly because the reward distribution of this game is very sparse that very limited sequence of actions can hit an effective reward. More intuitively, it can be explained as that the search space of Montezuma's Revenge appears to be very flat and there is very few information can be utilized to guide the search (i.e., to optimize the policy). On this basis, Montezuma's Revenge might be the most appropriate problem for assessing the exploration ability of RL methods. As both NCS and CCNCS can break the tie, the motivation of this paper that applying NCS to ERL can benefit the exploration ability is verified.

By comparing CCNCS with NCS, it suffices to show that the proposed NCS-friendly CC framework is able to facilitate NCS in the large-scale search space. That is, by consuming half less time than NCS, CCNCS can still significantly outperform NCS in almost all tested games. Contrarily, without using CC, the advantage of NCS over the three state-of-the-art algorithms is much weaker than that of CCNCS.

Table II The testing scores of compared algorithms on nine Atari games

| Methods | | CES | PPO | A3C | NCS | CCNCS | CCNCS |
|---|---|---|---|---|---|---|---|
| Time Budget | | 0.1B | 40M | 40M | 0.1B | 0.1B | 0.05B |
| Alien | Run 1 | 633.2 | 798.0 | 984.0 | 901.5 | 1211.6 | 1262.0 |
| | Run 2 | 640.5 | 800.7 | 1020.7 | 884.4 | 1231.4 | 1268.0 |
| | Run 3 | 416.0 | 301.0 | 328.3 | 1206.9 | 2658.9 | 1190.0 |
| | Average | 563.2 | 633.2 | 777.7 | 997.6 | **1700.6** | 1240.0 |
| Beamrider | Run 1 | 401.0 | 646.3 | 884.0 | 782.6 | 826.8 | 818.1 |
| | Run 2 | 508.2 | 573.2 | 513.5 | 648.7 | 908.4 | 682.3 |
| | Run 3 | 414.1 | 612.0 | 542.8 | 771.5 | 799.4 | 855.9 |
| | Average | 441.1 | 610.5 | 646.8 | 734.3 | **844.9** | 785.4 |
| Enduro | Run 1 | 6.2 | 501.6 | 0.0 | 11.3 | 13.3 | 11.0 |
| | Run 2 | 7.0 | 0.0 | 0.0 | 7.1 | 45.4 | 53.1 |
| | Run 3 | 8.1 | 400.0 | 0.0 | 7.8 | 9.8 | 6.6 |
| | Average | 7.1 | **300.5** | 0.0 | 8.7 | 22.8 | 23.6 |
| Freeway | Run 1 | 15.9 | 22.0 | 0.0 | 17.1 | 23.9 | 23.8 |
| | Run 2 | 12.7 | 11.3 | 0.0 | 19.3 | 24.7 | 24.4 |
| | Run 3 | 14.1 | 10.0 | 0.0 | 13 | 24.1 | 23.5 |
| | Average | 14.2 | 14.4 | 0.0 | 16.5 | **24.2** | 23.9 |
| Montezuma's Revenge | Run 1 | 0.0 | 0.0 | 0.0 | 6.5 | 72.0 | 57.1 |
| | Run 2 | 0.0 | 0.0 | 0.0 | 0 | 22.0 | 30.6 |
| | Run 3 | 0.0 | 0.0 | 0.0 | 2.0 | 0 | 0 |
| | Average | 0.0 | 0.0 | 0.0 | 2.8 | **31.3** | 29.2 |
| Pong | Run 1 | -19.9 | -19.8 | -21.0 | -9.8 | 4.8 | -9.9 |
| | Run 2 | -5.7 | -19.9 | -21.0 | -9.1 | 2.9 | 11.3 |
| | Run 3 | 6.3 | -20.0 | -21.0 | -9.6 | 4.0 | -10.8 |
| | Average | -6.4 | -19.9 | -21.0 | -9.5 | **3.9** | -3.5 |
| SpaceInvader | Run 1 | 340.0 | 551.5 | 500.0 | 1005 | 985.0 | 985.0 |
| | Run 2 | 275.0 | 468.2 | 944.0 | 800 | 1075.0 | 1045.0 |
| | Run 3 | 590.0 | 399.2 | 421.6 | 1035 | 1000.0 | 1000.0 |
| | Average | 401.0 | 473.0 | 621.9 | 946.0 | **1020.0** | 1010.0 |
| Venture | Run 1 | 420.0 | 0.0 | 0.0 | 527.0 | 657.0 | 663.3 |
| | Run 2 | 200.0 | 0.0 | 0.0 | 600.0 | 735.0 | 546.9 |
| | Run 3 | 400.0 | 0.0 | 0.0 | 418.0 | 748.0 | 791.8 |
| | Average | 340.0 | 0.0 | 0.0 | 515.0 | **713.0** | 667.3 |
| Bowling | Run 1 | 20.0 | 23.2 | 0.0 | 65.4 | 96.0 | 93.9 |
| | Run 2 | 92.5 | 23.9 | 0.0 | 70.2 | 78.1 | 47.0 |
| | Run 3 | 63.8 | 22.5 | 0.0 | 58.0 | 112.8 | 98.8 |
| | Average | 58.8 | 23.2 | 0.0 | 64.5 | **95.6** | 79.9 |
| Double Dunk | Run 1 | -3.3 | -3.5 | -2.0 | -1.5 | -0.6 | -0.4 |
| | Run 2 | -2.8 | -3.2 | -1.5 | -0.8 | 0.0 | 0.0 |
| | Run 3 | -4.1 | -3.0 | -2.8 | -1.7 | -0.1 | -0.4 |
| | Average | -3.4 | -3.2 | -2.1 | -1.3 | **-0.2** | -0.3 |

## Section V Conclusions

This paper studies how to effectively enable DRL approaches the exploration ability. By focusing on the emerging ERL techniques that employs EAs to optimize the parameters of the policies, this paper discusses that the recently presented NCS can be more appropriate to ERL due to its parallel exploration search behavior. To apply NCS to optimize millions of connection weights of neural policies, this paper employs the CC framework to scale up NCS for large-scale search space. The adopted CC framework is modified to specially fit NCS alike methods who have parallel exploration search behaviors. Extensive studies are conducted on 10 Atari games to verify the proposed method. Empirical results show that scaled-up NCS can perform more effectively than three state-of-the-art DRL methods with 50% less computational time, including both gradient-based ones and derivate-free ones. Furthermore, the powerful exploration ability enabled by NCS successfully helps the agent gain scores on the difficult Montezuma's Revenge game, where most existing methods cannot get scored due to the highly sparse reward distribution.